\documentclass[11pt]{article}

\usepackage[final]{acl}

\usepackage{times}
\usepackage{latexsym}

\usepackage[T1]{fontenc}
\usepackage[utf8]{inputenc}

\usepackage{microtype}

\usepackage{inconsolata}

\usepackage{graphicx}

\usepackage{adjustbox}
\usepackage{booktabs}
\usepackage{siunitx}
\usepackage{multirow}
\usepackage{adjustbox}
\usepackage{tcolorbox}
\usepackage{tablefootnote}

\usepackage{amsfonts}

\title{Selective Adversarial Attacks on LLM Benchmarks}

\author{Ivan Dubrovsky \\
  ITMO University\\
  Saint Petersburg \\
  \texttt{idubrovsky@itmo.ru} \\\And
  Anastasia Orlova \\
  ITMO University \\
  Saint Petersburg \\
  \texttt{orlova@scamt-itmo.ru} \\\And
  Illarion Iov \\
  ITMO University \\
  Saint Petersburg \\
  \texttt{illariov1809@gmail.com} \\\AND
  Nina Gubina \\
  ITMO University \\
  Saint Petersburg \\
  \texttt{gubina@scamt-itmo.ru} \\\And
  Irena Gureeva \\
  Applied AI Institute \\
  Moscow \\
  \texttt{irena-gureeva@mail.ru} \\\And
  Alexey Zaytsev \\
  Applied AI Institute \\
  Moscow \\
  \texttt{Likzet@gmail.com} 
  }

\begin{document}
\maketitle
\begin{abstract}
%Recent advancements in NLP, particularly the widespread adoption of LLMs, have brought to the forefront significant vulnerabilities. In this work we investigate adversarial attacks on benchmark datasets used to evaluate LLMs. Our aim was to design subtle perturbations of Massive Multitask Language Understanding (MMLU) benchmark questions that would selectively degrade or enhance the performance of a specific target LLM, while minimally affecting other models. Using TextAttack as our main framework, we successfully applied available attacks as well as integrated more recent custom attacks to degrade a target model's performance on MMLU by X\% while preserving the scores of other models. Conversely, we also achieved a Y\% improvement in a target model's performance, without impacting untargeted models by utilizing different set of attacks. These findings expose critical vulnerabilities in current LLM evaluation pipelines, highlighting the need for more robust benchmarks. Future work will focus on developing defensive mechanisms against such attacks.
Benchmarking outcomes increasingly govern trust, selection, and deployment of LLMs, yet these evaluations remain vulnerable to semantically equivalent adversarial perturbations. Prior work on adversarial robustness in NLP has emphasized text attacks that affect many models equally, leaving open the question of whether it is possible to selectively degrade or enhance performance while minimally affecting other models. We formalize this problem and study selective adversarial attacks on MMLU - a widely used benchmark designed to measure a language model’s broad general knowledge and reasoning ability across different subjects. Using canonical attacks integrated into TextAttack framework, we introduce a protocol for selectivity assessment, develop a custom constraint to increase selectivity of attacks and propose a surrogate-LLM pipeline that generates selective perturbations. Empirically, we find that selective adversarial attacks exist and can materially alter relative rankings, challenging the fairness, reproducibility, and transparency of leaderboard-driven evaluation. Our results motivate perturbation-aware reporting and robustness diagnostics for LLM evaluation and demonstrate that even subtle edits can shift comparative judgments.
\end{abstract}

\section{Introduction}

Large language models (LLMs) have rapidly become the cornerstone for a wide range of tasks, from general question answering and coding assistants \cite{wang2023review} to significant areas such as healthcare \cite{meng2024application} and education \cite{chu2025llm}. Due to the rapid integration of LLMs into many real-world applications, it is crucial to ensure their quality and reliability \cite{chang2024survey}. 

Demand for comprehensive models evaluation has led to the emergence of standardised benchmarks covering general natural language understanding, multitasking and reasoning \cite{hendrycks2020measuring, srivastava2023beyond}, as well as specialised knowledge \cite{rajpurkar2018know, hendrycks2021measuring}. The benchmarking results now play a decisive role in establishing trust, verifying capabilities and guiding implementation.

Therefore, the integrity of benchmark datasets is critical. Despite the careful design and continuous refinement of widely used benchmarks \cite{wang2024mmlu, gema2024we}, LLMs sensitivity to input perturbations remains an issue \cite{sclar2023quantifying, biswas2025universal}. Subtle adversarial manipulations – small edits that change model behavior without altering perceived meaning – can significantly inflate or deflate performance metrics \cite{hendrycks2021measuring, clark2018think}. Such attacks undermine fair comparison among competing models and raise concerns about reproducibility and transparency of published results.

Over the past decade, numerous research papers have been published on the generic robustness of LLMs to attacks through perturbations at the character-, -word-, and sentence-level, or universal trigger \cite{ebrahimi2017hotflip, jin2020bert, zhang2021crafting}. In order to standardize the application of classical adversarial attacks, frameworks have been developed \cite{morris2020textattack, zeng2020openattack, zhu2023promptbench} that unify goal functions, constraints, transformations, and search methods to simplify the development of new tools and the application of existing ones. Meanwhile, most research focuses on non-selective degradation, meaning perturbations that reduce the performance of many models.

In contrast, selective attacks that degrade the performance of the target model without affecting others remain largely unexplored. This scenario is particularly relevant in competitive settings, where even small differences in evaluation can influence deployment decisions and public opinion. Our work bridges this gap by investigating perturbations of commonly used benchmark datasets that cause the target LLM to perform worse (or better) than non-target models. Our main contributions can be summarized as follows:

\begin{itemize}
    \item To the best of our knowledge, we are the first to formulate the problem of selectivity in adversarial attacks and conduct a systematic comparison of target and non-target effects on several LLMs using TextAttack.
    \item We propose a white-box attack pipeline based on a surrogate model to generate selective perturbations, enabling attacks without access to target internals.
    \item We empirically show that the attack degrades only the target, leaving non-targets intact across setups, including same-family models.
    \item We publish perturbed open datasets constructed under the proposed protocol to facilitate robust, manipulation-resistant evaluation.
\end{itemize}

\section{Related Works}

\subsection{LLM Benchmarks}
As the result of extensive research, general and domain-specific tests were developed, which became the standard for comparing language models. Early comprehensive datasets, such as GLUE \cite{wang2018glue} and SuperGLUE \cite{wang2019superglue}, catalyzed standardized evaluation of general language understanding, while domain-specific resources targeted specific reasoning skills. For example, specialized benchmarks such as SQuAD \cite{rajpurkar2016squad, rajpurkar2018know} for reading comprehension, HellaSwag \cite{zellers2019hellaswag} and ARC \cite{clark2018think} for common sense and reasoning, GSM8K \cite{cobbe2021training} and MATH \cite{hendrycks2021measuring} for mathematical knowledge have been developed. As the capabilities of models grew, the community introduced benchmarks focused on broadness and logical reasoning: MMLU \cite{hendrycks2020measuring} for academic and professional knowledge across 57 subjects, BIG-bench \cite{srivastava2023beyond} for a variety of tasks beyond the capabilities of language models at the time, and HELM for evaluation across a wide range of scenarios and multiple performance metrics. More specialized benchmarks continue to be developed, such as GPQA \cite{rein2024gpqa} and Humanity’s Last Exam \cite{phan2025humanity}, which were created by domain experts to remain challenging even for state-of-the-art LLMs. In addition, recent efforts to refine the MMLU dataset \cite{wang2024mmlu, gema2024we} aim to mitigate sensitivity to prompts, noise in datasets, and errors - problems that can complicate comparisons between different models.

\subsection{Adversarial Text Attacks}
Adversarial robustness in NLP has been studied at multiple levels of granularity and with diverse optimization strategies. Character-level attacks such as HotFlip \cite{ebrahimi2017hotflip} and DeepWordBug \cite{gao2018black} exploit gradient or heuristic signals to induce typos and visually confusable substitutions. Word-level approaches – including TextBugger \cite{li2018textbugger}, PWWS \cite{ren2019generating}, TextFooler \cite{jin2020bert}, and BERT-Attack \cite{li2020bert} – search for lexically minimal edits under constraints such as semantic similarity and textual consistency. Sentence-level perturbations leverage paraphrasing \cite{ribeiro2018semantically}, back-translation \cite{wallace2020imitation, zhang2021crafting}, or distraction-style edits \cite{qi2021mind} to alter model decisions. Beyond instance-specific edits, universal triggers (short token sequences prepended or appended to inputs) have been shown to induce consistent failure modes across many examples and tasks \cite{wallace2019universal, xu2024linkprompt}. Frameworks such as TextAttack \cite{morris2020textattack}, OpenAttack \cite{zeng2020openattack}, and PromptBench \cite{zhu2023promptbench} have unified goal functions, constraints, transformations, and search strategies, enabling reproducible comparisons and rapid integration of novel adversarial attacks. Furthermore, adversarial attacks are often categorized by the level of access to the model into white-box (full access to parameters and gradients), gray-box (partial knowledge), and black-box (query-only access) scenario \cite{ma2025safety}. Recent work has focused on developing black-box attacks at different granularities \cite{rocamora2024revisiting, liu2024hygloadattack, formento2025confidence} that do not access LLM internals and either operate on hard-labels (decision-only) or leverage soft-labels (confidence scores/logits). Despite this progress, most studies evaluate standard metrics that emphasize non-selective degradation rather than differential impact across models competing on the same benchmark \cite{qiu2022adversarial, goyal2023survey}. As a result, the literature offers limited guidance on constructing perturbations that reliably change performance of a target model while leaving non-target models mostly unaffected.

\subsection{Selectivity of Adversarial Attacks}
The process of degrading (or enhancing) a target model’s benchmark performance while minimally affecting non-targets, namely selectivity, is closely connected to transferability and benchmarking. In vision and classical NLP robustness, transfer studies show that some adversarial examples are model-specific while others generalize broadly \cite{zheng2023large, alzahrani2024benchmarks}, but this property has rarely been operationalized as an explicit objective in text attacks. Closest topics include: (i) analyses of cross-model transfer for word- and character-level attacks (which implicitly reveal non-transferable and potentially selective examples) \cite{sclar2023quantifying, nalbandyan2025score}, (ii) adversarial data collection protocols (DynaBench) where failures are found against a current leading model and later tested on new models \cite{kiela2021dynabench}, and (iii) safety or jailbreak literature demonstrating model-specific prompt suffixes and exploits evidence that targeted, architecture or training-data dependent vulnerabilities exist \cite{wang2024attngcg, biswas2025universal}. However, these researches typically do not evaluate rank instability on competitive leaderboards, nor do they provide a systematic protocol to seek perturbations that maximize a target/non-target gap under semantic constraints. Our work makes this notion explicit: we formalize selectivity as a controlled difference in performance between a chosen target LLM and a comparison set.

\begin{figure*}
  \centering
  \includegraphics[width=1\textwidth]{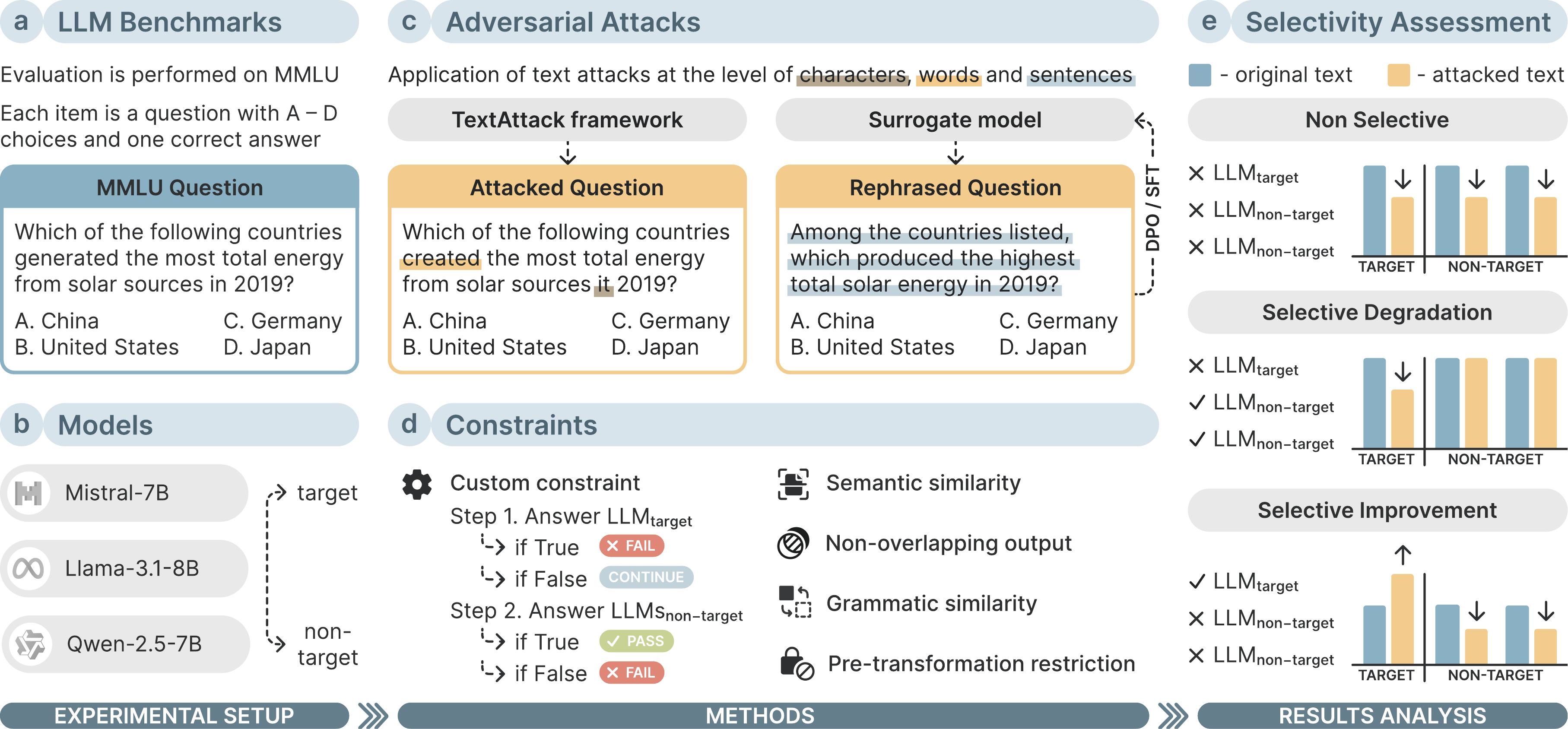}
  \caption{Overview of selective adversarial evaluation. (a) LLM benchmarks: start from standard multiple-choice items from MMLU dataset. (b) Constraints: edits are subject to constraints from the categories semantic similarity, grammatical similarity, non-overlapping output, pre-transformation restriction, and a custom constraint for implementing selectivity in the TextAttack framework. (c) Adversarial attacks: using TextAttack, apply character-, word-, and sentence-level transformations to the question or instruction. A surrogate generator proposes paraphrases; candidates are filtered by the constraints and can be improved via SFT/DPO to preferentially degrade a chosen model. (d) Models: three open LLMs (Mistral-7B, Llama-3.1-8B, Qwen-2.5-7B); one is designated as the target and the others as non-targets. (f) Selectivity assessment: compare accuracy on original and perturbed items and categorize outcomes as non-selective (all models change similarly), selective degradation (only the target degrades), or selective improvement (only the target improves).}
  \label{fig_abstr}
\end{figure*}

\section{Methods}

\subsection{Problem Setting}

We investigate the robustness of LLMs under targeted perturbations of evaluation benchmarks. Specifically, we aim to construct adversarial versions of the Massive Multitask Language Understanding (MMLU) benchmark questions that reduce the performance of a chosen \textit{target} model, while maintaining the performance of other reference models.

Let \( Q \) be a question from the original benchmark, with answer choices \( A = \{A_1, A_2, \ldots, A_n\} \), and correct index \( y \in \{1, \ldots, n\} \). 
Given a target model \( M_t \) and a set of reference models \( \{M_1, \ldots, M_k\} \), our objective is to produce a perturbed question \( Q' \) such that:
\begin{itemize}
    \item \( M_t(Q') \ne y \) \hfill (target model fails),
    \item \( M_t(Q) = y \) \hfill (target model succeeds initially),
    \item \( M_i(Q') = M_i(Q) = y \quad \forall i \in [1,k] \) \hfill (reference models unaffected).
\end{itemize}

\subsection{Validation protocol}

\paragraph{Dataset}
Experiments use the MMLU benchmark (License: MIT License) \cite{hendrycks2020measuring, hendrycks2021ethics}, which covers 57 academic and professional subjects across humanities, social sciences, STEM, and other knowledge areas. Each item consists of a natural-language question and four options labeled A through D with a single correct label. We utilize the development split (\texttt{dev}) of 285 samples due to the computational cost of processing the full benchmark.
% We evaluate on the public test split in the standard multiple-choice setting. Unless specified, the full set of items is included.

\paragraph{Models}
We use \texttt{Qwen2-7B} (Apache license 2.0) \cite{yang2024qwen2}, \texttt{Llama-3.1-8B} (Llama 3.1 Community License Agreement) \cite{grattafiori2024llama}, and \texttt{Mistral-7B} (Apache license 2.0) \cite{jiang2023mistral} because they are widely adopted open models with comparable parameter scales, diverse training corpora and architectures, and strong baseline performance on MMLU, which makes them suitable for studying selective robustness. For each experimental condition, one model is designated as \(M_t\) and the remaining two as \(\mathcal{M}_r\). Decoding is deterministic with temperature set to 0 and nucleus sampling disabled. The generation budget is capped at one new token to elicit a single-letter answer. Outputs are normalized to \(\{A,B,C,D\}\) by taking the first valid letter. Responses without a valid letter are scored as incorrect.

In addition, we conducted supplementary experiments across model families with different parameter counts to examine the relationship between scale and robustness. These include the \texttt{Llama‑3.2} series (1B, 3B, and 11B‑Vision), as well as the \texttt{Qwen2} series (1.5B and 7B).

\paragraph{Metrics}
To quantify both overall performance and the impact of perturbations, we report:
\begin{itemize}
    \item \textbf{Accuracy} on original \(S_{\text{base}}\) and perturbed \(S_{\text{attack}}\) items.
    \item \textbf{Manipulation Magnitude (MM)}, the absolute change in accuracy,
    \begin{equation}
    \text{$\Delta$}=S_{\text{attack}}-S_{\text{base}}
    \end{equation}
\end{itemize}

% Unless noted, metrics are computed per subject and macro-averaged across subjects. Confidence intervals are obtained by three repeated launches of each LLM on the same task configuration.

\subsection{TextAttack framework}

\begin{table}[h]
\centering
\caption{Attack recipes with different levels used for experiments from the TextAttack framework.}
\label{tab:attack_recipes}
\begin{adjustbox}{max width=1\columnwidth}
\small
\begin{tabular}{@{} l l l @{}}
\toprule
\multicolumn{1}{c}{Attack recipe} &
\multicolumn{1}{c}{Attack level} &
\multicolumn{1}{c}{Reference} \\
\midrule
\texttt{hotflip}        & \multirow{2}{*}{Character} &  \cite{ebrahimi2017hotflip} \\
\texttt{deepwordbug}    &                             & \cite{gao2018black} \\
\midrule
\texttt{textbugger}             & \multirow{14}{*}{Word} & \cite{li2018textbugger} \\
\texttt{pruthi}                 &                        & \cite{pruthi2019combating} \\
\texttt{kuleshov}               &                        & \cite{kuleshov2018adversarial} \\
\texttt{textfooler}             &                        & \cite{jin2020bert} \\
\texttt{pwws}                   &                        & \cite{ren2019generating} \\
\texttt{bae}                    &                        & \cite{garg2020bae}\\
\texttt{bert-attack}            &                        & \cite{li2020bert} \\
\texttt{iga}                    &                        & \cite{wang2019natural} \\
\texttt{genetic-algo}      &                        & \cite{alzantot2018generating} \\
\texttt{fast-genetic-algo} &                        & \cite{jia2019certified} \\
\texttt{pso}                    &                        & \cite{zang2019word} \\
\texttt{clare}                  &                        & \cite{li2020contextualized} \\
\texttt{checklist}              &                        & \cite{ribeiro2020beyond} \\
\texttt{a2t}                    &                        & \cite{yoo2021towards} \\
\midrule
\texttt{input-reduction} & Sentence & \cite{feng2018pathologies} \\
\bottomrule
\end{tabular}
\end{adjustbox}
\end{table}

Attacks are implemented with \texttt{TextAttack} (License: MIT License) \cite{morris2020textattack}. The goal function is \texttt{Untargeted Classification}, which aims to induce any label other than the correct one for the target model. The search method is \texttt{GreedySearch} under the semantic and syntactic constraints described below. 

All attacks (\autoref{tab:attack_recipes}) were conducted under a full white‑box setting. Each model was deployed locally, and the internal probability distributions were directly accessible. After every inference step, we extracted the raw output logits corresponding to the four multiple‑choice options \(\{A,B,C,D\}\) and converted them into probabilities by applying a softmax transformation. This allowed us to compute model confidence for each answer option and use these probabilities to guide adversarial search more precisely.

\paragraph{Constraints}
We allow modifications only to the system instruction and the question text while keeping answer options unchanged. 
% A stop-list prevents edits to stop words and to the option letters A, B, C, and D. This preserves the integrity of the multiple-choice format and reduces trivial failure modes.

Attacks proceed iteratively over candidate edits proposed by the recipe's search strategy and terminate upon success or exhaustion of candidates. All edits are governed by the following four categories of constraints (the specific set of constraints was set depending on the attack used):
\begin{itemize}
  \item \textbf{Semantic similarity}. Preserve the original meaning at the sentence.
  \item \textbf{Grammatic similarity}. Maintain grammatical role of substitutions.
  \item \textbf{Non-overlapping output}. Keep perturbations small and non-redundant.
  \item \textbf{Pre-transformation restriction}. Limit what can be edited before search begins.
\end{itemize}

In addition to the default constraint described above, we introduced a custom variant designed to make the attack selective to a predefined target model. The constraint checks each version of the perturbed question to ensure that the target model gives incorrect answer while reference models produce the right ones. Under this setup, the target model’s accuracy should decline, whereas the accuracies of the reference models are expected to remain the same or improve. This modification guides the search algorithm toward transformations that specifically impair the target model’s performance, thereby amplifying selective effects.

\paragraph{Experimental design}
We conducted three distinct types of experiments.

(1) The first experiment compared the selectivity of attacks across three models of comparable parameter scale: \texttt{Qwen2‑7B}, \texttt{Llama‑3.1‑8B}, and \texttt{Mistral‑7B}. In each run one model was designated as the target and the remaining two as references. We computed accuracy before and after the attack for each model and defined the attack effect as their difference. Selectivity was achieved when the manipulation magnitude between the target and reference deltas exceeded \textit{0.10}.

(2) The second experiment analyzed attacks within a single model family differing in parameter count. Metrics were computed as above, but selective behavior was defined as the case in which the smaller model outperformed the larger one after the attack. The larger model in each family always served as the target. We evaluated three such families: (\texttt{Llama‑3.2‑1B}, \texttt{Llama‑3.2‑3B}, \texttt{Llama‑3.2‑11B‑Vision}), (\texttt{Llama‑3.2‑1B}, \texttt{Llama‑3.2‑3B}), and (\texttt{Qwen2‑1.5B}, \texttt{Qwen2‑7B}).

(3) The third experiment investigated transferability and surrogate‑based attacks described in the \textit{Surrogate model} section; its results are reported jointly with those of the surrogate experiments.

For the first two experiments we employed two attack recipes that showed the highest potential for inducing selective degradation in preliminary screening: the word‑level \textbf{BAE (BERT‑Based Adversarial Examples)} \cite{garg2020bae} and the character‑level \textbf{DeepWordBug} \cite{gao2018black}. Both attacks were selected after evaluating all methods listed in \autoref{tab:attack_recipes}. In the first two experiments, the metrics were calculated both using the original attack recipe and with a custom constraint (described in the Constraints subsection) responsible for the selectivity of the attack. These two setups allowed us to evaluate how the default and selective conditions influence the success rate and robustness of attacks.

\subsection{Surrogate model}

\paragraph{Dataset construction for fine-tuning}
To generate selective perturbations without accessing internals of \(M_t\), a surrogate generator \(M_s\) produces paraphrases \(\{Q'_1,\dots,Q'_m\}\) for each original item \(Q\). Each paraphrase is evaluated by \(M_t\) and by \(\mathcal{M}_r\). We score paraphrases using a weighted cost function with two components: misclassification (maximized when the target model is incorrect) and consistency (maximized when reference models preserve their original responses, regardless of correctness). From this scoring we retain two classes of samples:
\begin{itemize}
\item \textbf{Best:} highest-scoring paraphrases (strong misclassification signal on \(M_t\) while preserving reference-model responses).
\item \textbf{Worst:} lowest-scoring paraphrases (little or no misclassification signal on \(M_t\), and/or perturb reference-model responses).
\end{itemize}
This assessment step ensures the modified questions preserve semantic content while selectively changing the performance of the target model. At the same time, the best paraphrases already satisfy the criteria for a selective attack, enabling attacks without additional surrogate-model training. In the results, we refer to this sampling-based method as \emph{paraphrase sampling}.

\paragraph{Training objectives}
After initial paraphrased question sampling, we train the surrogate model to improve its generation of selective-attack questions. We experiment with three learning strategies:

\begin{enumerate}
\item \textbf{Supervised Fine-Tuning (SFT).} We fine-tune the surrogate model on the \textbf{best} paraphrase samples, i.e., those that induce target-model failures while preserving reference-model behavior.
\item \textbf{Direct Preference Optimization (DPO).} \cite{rafailov2023direct} We construct a preference dataset using pairs drawn from the retained samples: the \textit{chosen} item is the \textbf{best} sample and the \textit{rejected} item is the corresponding \textbf{worst} sample. This encourages the model to prefer constructions that maximize the adversarial gap between \(M_t\) and \(M_r\).
\end{enumerate}

The DPO objective therefore pushes the surrogate to favor prompt formulations that widen the adversarial gap, while SFT directly fits the model to successful adversarial paraphrases.

\paragraph{Inference with the surrogate model}
At test time, \(M_s\) generates a small batch of candidates per item using low-temperature sampling. Candidates are filtered by the same semantic and syntactic constrains as above and are then evaluated against \(M_t\) and \(\mathcal{M}_r\). If the strict selectivity criterion is not met, we select the candidate that maximizes the difference between baseline and attack accuracy for the target model under all constraints.

\paragraph{Cycle training}
After generating new samples, they can be evaluated with the same cost function and then employed in another surrogate model training cycle further improving the results.

\section{Results}
\subsection{Selective attacks without the custom constraint}

In the baseline configuration, attacks were executed under the default constraint, without any additional selectivity objectives. The obtained results (Table \ref{tab:exp1_combined_results}a) show that the inherent selectivity of the chosen attack methods is relatively low. Both BAE and DeepWordBug attacks caused moderate performance degradation across models, but the average differences between target and reference models rarely exceeded 0.05 in accuracy. This means that without explicitly guiding the perturbation process through a custom constraint, both attacks tend to affect all models in a comparable way rather than producing highly selective outcomes.

The accuracy drops produced by DeepWordBug were generally smaller than those of the BAE attack. The lower selectivity of the character‑level method reflects the fact that modern language models are considerably more resilient to individual character variations and typographical noise. Minor symbol‑level manipulations are likely handled in tokenization and normalized during inference. In contrast, word‑level replacements, as in the BAE Attack, can subtly alter meaning or discourse context, producing more substantial cognitive shifts in model reasoning and therefore larger effects on accuracy.

\begin{table}[tbh!]
\centering
\caption{Results for the original (a) and custom constraint (b) implementations of the attack recipes. Subscripts indicate the change ($\Delta$) from the original performance, with arrows denoting direction (↑ improvement, ↓ degradation). For the target model (\dag), lower values indicate better selectivity; for all other models, minimal or no change is preferable.}
\label{tab:exp1_combined_results}

\begin{adjustbox}{width=\columnwidth}
\begin{tabular}{@{}lccc@{}}
\toprule
\multirow{2}{*}{Target model} & \multirow{2}{*}{Before attack} & \multicolumn{2}{c}{After attack $_{\Delta}$} \\ 
\cmidrule(l){3-4}
& & BAE & DeepWordBug \\ 
\midrule
\multicolumn{4}{@{}l}{\textbf{(a) Baseline}} \\ 
\midrule
Mistral-7B $\dag$             & 0.59 & 0.47$_{-0.12\downarrow}$ & 0.47$_{-0.12\downarrow}$ \\
Qwen2-7B                      & 0.74 & 0.68$_{-0.06\downarrow}$ & 0.72$_{-0.02\downarrow}$ \\
Llama-3.1-8B                  & 0.69 & 0.64$_{-0.05\downarrow}$ & 0.65$_{-0.04\downarrow}$ \\ \midrule
Mistral-7B                    & 0.59 & 0.51$_{-0.08\downarrow}$ & 0.55$_{-0.04\downarrow}$ \\
Qwen2-7B $\dag$               & 0.74 & 0.52$_{-0.22\downarrow}$ & 0.59$_{-0.15\downarrow}$ \\
Llama-3.1-8B                  & 0.69 & 0.60$_{-0.09\downarrow}$ & 0.64$_{-0.05\downarrow}$ \\ \midrule
Mistral-7B                    & 0.59 & 0.53$_{-0.06\downarrow}$ & 0.58$_{-0.01\downarrow}$ \\
Qwen2-7B                      & 0.74 & 0.69$_{-0.05\downarrow}$ & 0.74$_{-0.00}$\hspace{0.4em} \\
Llama-3.1-8B $\dag$           & 0.69 & 0.53$_{-0.16\downarrow}$ & 0.61$_{-0.08\downarrow}$ \\ 
\midrule
\multicolumn{4}{@{}l}{\textbf{(b) With a custom selective constraint}} \\ 
\midrule
Mistral-7B $\dag$             & 0.59 & 0.46$_{-0.13\downarrow}$ & 0.47$_{-0.12\downarrow}$ \\
Qwen2-7B                      & 0.74 & 0.74$_{-0.00}$\hspace{0.4em} & 0.75$_{+0.01\uparrow}$ \\
Llama-3.1-8B                  & 0.69 & 0.71$_{+0.02\uparrow}$ & 0.69$_{-0.00}$\hspace{0.4em} \\ 
\midrule
Mistral-7B                    & 0.59 & 0.51$_{-0.08\downarrow}$ & \hspace{0.4em}0.40$_{-0.02\downarrow}$* \\ 
Qwen2-7B $\dag$               & 0.74 & 0.47$_{-0.27\downarrow}$ & \hspace{0.4em}0.38$_{-0.36\downarrow}$* \\ 
Llama-3.1-8B                  & 0.69 & \hspace{0.4em}0.40$_{-0.29\downarrow}$* & \hspace{0.4em}0.40$_{-0.29\downarrow}$* \\ 
\midrule
Mistral-7B                    & 0.59 & \hspace{0.4em}0.44$_{-0.15\downarrow}$* & 0.59$_{-0.00}$\hspace{0.4em} \\ 
Qwen2-7B                      & 0.74 & \hspace{0.4em}0.40$_{-0.34\downarrow}$* & 0.75$_{+0.01\uparrow}$ \\ 
Llama-3.1-8B $\dag$           & 0.69 & \hspace{0.4em}0.32$_{-0.37\downarrow}$* & 0.63$_{-0.06\downarrow}$ \\ 
\bottomrule
\multicolumn{4}{@{}l}{\footnotesize * results obtained on college chemistry subset; results on the full dev split} \\ [-0.35ex]
\multicolumn{4}{@{}l}{\footnotesize \hspace{0.5em} will be added in camera-ready version.}\\
\end{tabular}
\end{adjustbox}

\end{table}

\subsection{Experiments with the custom constraint}

Introducing the custom constraint led to clearer and more consistent selective behavior (Table \ref{tab:exp1_combined_results}b). In this setup, the constraint was designed to force the post‑attack accuracy of the target model toward zero while preserving the baseline performance of reference models. Under this condition, the same BAE and DeepWordBug recipes revealed a much stronger divergence between models: the target model exhibited a substantial accuracy decline, whereas the metrics of non‑target models remained almost unchanged.

In particular, the BAE attack demonstrated the highest level of selectivity. For several target configurations, the difference in delta between the target and the reference models exceeded 0.10. The strongest selective effect appeared when \texttt{Qwen2‑7B} served as the target model, with an average decline of –0.14 to –0.26 across runs with and without custom constraint. This suggests that Qwen models respond more sensitively to meaning‑modifying word substitutions, which may stem from differences in their training corpora, tokenization, or alignment strategies compared with the Llama and Mistral families. Because Qwen models consistently produced the most selective results, it was chosen as the principal target model for subsequent experiments with sentence‑level attacks in the surrogate‑model framework.

Conversely, the character‑level DeepWordBug Attack remained less effective even under the custom constraint. This again indicates that current LLMs are comparatively robust to single‑symbol substitutions but remain more vulnerable to semantically meaningful word‑level changes.

% \begin{table}[tbh!]
% \centering
% \caption{Results for the implementations of the attack recipes with a custom selective constraint. Subscripts indicate the change ($\Delta$) from the original performance, with arrows denoting direction (↑ improvement, ↓ degradation). For the target model (\dag), lower values indicate better selectivity; for all other models, minimal or no change is preferable.}
% \label{tab:exp1_selective_results}
% \begin{adjustbox}{width=\columnwidth}
% \begin{tabular}{@{}lccc@{}}
% \toprule
% \multirow{2}{*}{Target model} & \multirow{2}{*}{Before attack} & \multicolumn{2}{c}{After attack $_{\Delta}$} \\ 
% \cmidrule(l){3-4}
%                               &                                & BAE & DeepWordBug \\ 
% \midrule
% Mistral-7B $\dag$             & 0.59 & 0.46$_{-0.13\downarrow}$ & 0.47$_{-0.12\downarrow}$ \\
% Qwen2-7B                      & 0.74 & 0.74$_{-0.00}$\hspace{0.4em} & 0.75$_{+0.01\uparrow}$ \\
% Llama-3.1-8B                  & 0.69 & 0.71$_{+0.02\uparrow}$ & 0.69$_{-0.00}$\hspace{0.4em} \\ 
% \midrule
% Mistral-7B                    & 0.59 & 0.51$_{-0.08\downarrow}$ & -- \\ 
% Qwen2-7B $\dag$               & 0.74 & 0.47$_{-0.27\downarrow}$ & -- \\ 
% Llama-3.1-8B                  & 0.69 & -- & -- \\ 
% \midrule
% Mistral-7B                    & 0.59 & -- & 0.59$_{-0.00}$\hspace{0.4em} \\ 
% Qwen2-7B                      & 0.74 & -- & 0.75$_{+0.01\uparrow}$ \\ 
% Llama-3.1-8B $\dag$           & 0.69 & -- & 0.63$_{-0.06\downarrow}$ \\ 
% \bottomrule
% \end{tabular}
% \end{adjustbox}
% \end{table}

\subsection{Model‑family experiments}

The second set of experiments tested attacks within homogeneous model families differing mainly in size, isolating the effect of scale. Results (Tables \ref{tab:exp2_combined_results}) showed similar trends: in the BAE Attack, smaller models often matched or outperformed larger ones, while DeepWordBug remained largely ineffective.

An intriguing observation is that on the evaluated MMLU subset, smaller models such as \texttt{Llama‑3.2‑1B} and \texttt{Llama‑3.2‑3B} displayed baseline accuracies nearly identical to those of larger versions, which further amplified the visible effects of selective perturbations. In several runs the largest model in the family, for example \texttt{Llama‑3.2‑11B‑Vision}, experienced a substantially stronger drop in post‑attack accuracy compared with smaller siblings. These results reinforce the idea that parameter scaling alone does not guarantee robustness and, under certain perturbation patterns, larger models may be disproportionately sensitive.

Overall, across all experiments, the custom‑constraint configuration proved crucial for eliciting selective effects, the BAE attack was the most effective in inducing them, and \texttt{Qwen2‑7B} emerged as the most distinctively vulnerable target, thus serving as the primary candidate for subsequent evaluations with more complex sentence‑level perturbations and surrogate‑based attack generation.

\begin{table}[h!]
\centering
\caption{Benchmark results for selective attacks on different size models of the same families. Subscripts indicate the change ($\Delta$) from the original performance, with arrows denoting direction (↑ improvement, ↓ degradation). For the target model (\dag), lower values indicate better selectivity; for all other models, minimal or no change is preferable.}
\label{tab:exp2_combined_results}

\begin{adjustbox}{width=\columnwidth}
\begin{tabular}{@{}lccc@{}}
\toprule
\multirow{2}{*}{Target model} & \multirow{2}{*}{Before attack} & \multicolumn{2}{c}{After attack $_{\Delta}$} \\ 
\cmidrule(l){3-4}
& & BAE & DeepWordBug \\ 
\midrule
\multicolumn{4}{@{}l}{\textbf{(a) Baseline}} \\ 
\midrule
Llama-3.2-1B                  & 0.45 & 0.42$_{-0.03\downarrow}$ & 0.41$_{-0.04\downarrow}$ \\
Llama-3.2-3B                  & 0.59 & 0.51$_{-0.08\downarrow}$ & 0.56$_{-0.03\downarrow}$ \\
Llama-3.2-11B $\dag$   & 0.69 & 0.52$_{-0.17\downarrow}$ & 0.61$_{-0.08\downarrow}$ \\ \midrule
Llama-3.2-1B                  & 0.45 & 0.40$_{-0.05\downarrow}$ & 0.43$_{-0.02\downarrow}$ \\
Llama-3.2-3B $\dag$           & 0.59 & 0.42$_{-0.17\downarrow}$ & 0.51$_{-0.08\downarrow}$ \\ \midrule
Qwen2-1.5B                    & 0.58 & 0.51$_{-0.07\downarrow}$ & 0.53$_{-0.05\downarrow}$ \\
Qwen2-7B $\dag$               & 0.74 & 0.69$_{-0.05\downarrow}$ & 0.74$_{-0.00}$\hspace{0.4em} \\
\midrule
\multicolumn{4}{@{}l}{\textbf{(b) With a custom selective constraint}} \\ 
\midrule
Llama-3.2-1B                  & 0.45 & 0.45$_{-0.00}$\hspace{0.4em} & 0.44$_{-0.01\downarrow}$ \\
Llama-3.2-3B                  & 0.59 & 0.60$_{+0.01\uparrow}$ & 0.59$_{-0.00}$\hspace{0.4em} \\
Llama-3.2-11B $\dag$   & 0.69 & 0.59$_{-0.10\downarrow}$ & 0.67$_{-0.02\downarrow}$ \\ \midrule
Llama-3.2-1B                  & 0.45 & 0.46$_{+0.01\uparrow}$ & 0.47$_{+0.02\uparrow}$ \\
Llama-3.2-3B $\dag$           & 0.59 & 0.48$_{-0.11\downarrow}$ & 0.53$_{-0.06\downarrow}$ \\ \midrule
Qwen2-1.5B                    & 0.58 & 0.62$_{+0.04\uparrow}$ & 0.59$_{+0.01\uparrow}$ \\
Qwen2-7B $\dag$               & 0.74 & 0.58$_{-0.16\downarrow}$ & 0.68$_{-0.06\downarrow}$ \\
\bottomrule
\end{tabular}
\end{adjustbox}

\end{table}

\subsection{Paraphrase selective adversarial attacks using a surrogate model}

% % Initial DPO Results
% \begin{table}[h!]
% \centering
% \caption{Benchmark results. For the target model Qwen3\dag, lower is better; for others, no change is better.}

% \label{tab:results}
% \small
% \begin{tabular}{l
%                 S % Initial
%                 S % Random
%                 S % Δ Random
%                 }
% \toprule
% & {Initial} & \multicolumn{2}{c}{Random Search} \\
% \cmidrule(lr){2-2} \cmidrule(lr){3-4}
% {Model} & {Result} & {Result} & {$\Delta$} \\
% \midrule
% Qwen3\dag   & 0.73 & 0.67 & -0.06\\
% Mistral & 0.53 & 0.51 & -0.02 \\
% LLaMA   & 0.64 & 0.62 &  -0.02  \\
% \bottomrule
% \end{tabular}

% \vspace{0.5em} % small space between "floors"

% \begin{tabular}{l
%                 S % DPO
%                 S % Δ DPO
%                 S % SFT
%                 S
%                 }
% \toprule
% & \multicolumn{2}{c}{SFT} & \multicolumn{2}{c}{DPO} \\
% \cmidrule(lr){2-3} \cmidrule(lr){4-5}
% {Model} & {Result} & {$\Delta$} & {Result} & {$\Delta$} \\
% \midrule
% Qwen3\dag   & 0.68 & -0.05 & 0.58 & -0.10 \\
% Mistral & 0.51 & -0.02 & 0.47 & -0.06 \\
% LLaMA   & 0.63 &  -0.01 & 0.57 & -0.07 \\
% \bottomrule
% \end{tabular}
% \end{table}
\paragraph{Minimisation of target model score}

The results for one to three training cycles are shown in Table \ref{tab:paraphrase_results_combined}. We compare the training results of SFT/DPO to the base benchmark scores and paraphrase sampling from non-trained surrogate model.

\begin{table*}[t]
\centering
\caption{Benchmark results across training iterations for \textbf{minimizing} target model score in surrogate model method. For the target model Qwen2.5-7B\dag, lower is better; for others, no change is better. $\Delta$ represents change from initial baseline; $\uparrow/\downarrow$ indicate direction. The best result in each category is highlighted in bold, the second best result is underlined.}
\label{tab:paraphrase_results_combined}
\small
\begin{adjustbox}{width=\textwidth}
\begin{tabular}{l c c c c c c c c}
\toprule
& \multirow{2}{*}{\shortstack{Before\\Attack}} 
& \multirow{2}{*}{\shortstack{Paraphrase\\Sampling}} 
& \multicolumn{3}{c}{SFT} 
& \multicolumn{3}{c}{DPO} \\
\cmidrule(lr){4-6} \cmidrule(lr){7-9}
Model &  &  & Iter 1 & Iter 2 & Iter 3 & Iter 1 & Iter 2 & Iter 3 \\
\midrule
Qwen2.5-7B\dag & 0.81 & 0.75$_{-0.06\downarrow}$ & 0.74$_{-0.07\downarrow}$ & 0.74$_{-0.07\downarrow}$ & 0.74$_{-0.07\downarrow}$ & 0.73$_{-0.08\downarrow}$ & \underline{0.72}$_{-0.09\downarrow}$ & \textbf{0.71}$_{-0.10\downarrow}$ \\
Mistral-7B & 0.52 & 0.48$_{-0.04\downarrow}$ & 0.48$_{-0.04\downarrow}$ & 0.49$_{-0.03\downarrow}$ & 0.50$_{-0.02\downarrow}$ & 0.50$_{-0.02\downarrow}$ & \textbf{0.51}$_{-0.01\downarrow}$ & \textbf{0.51}$_{-0.01\downarrow}$ \\
Llama-3.1-8B & 0.64 & 0.59$_{-0.05\downarrow}$ & \underline{0.60}$_{-0.04\downarrow}$ & \underline{0.60}$_{-0.04\downarrow}$ & \underline{0.60}$_{-0.04\downarrow}$ & \underline{0.60}$_{-0.04\downarrow}$ & \underline{0.60}$_{-0.04\downarrow}$ & \textbf{0.62}$_{-0.02\downarrow}$ \\
\bottomrule
\end{tabular}
\end{adjustbox}
\end{table*}

\begin{table*}[t]
\centering
\caption{Benchmark results across training iterations for \textbf{maximizing }target model score in surrogate model method. For the target model Qwen2.5-7B\dag, higher is better; for others, no change is better. $\Delta$ represents change from initial baseline; $\uparrow/\downarrow$ indicate direction. The best result in each category is highlighted in bold, the second best result is underlined.}
\label{tab:paraphrase_results_higher_target}
\small
\begin{adjustbox}{width=\textwidth}
\begin{tabular}{l c c c c c c c c}
\toprule
& \multirow{2}{*}{\shortstack{Before\\Attack}} 
& \multirow{2}{*}{\shortstack{Paraphrase\\Sampling}} 
& \multicolumn{3}{c}{SFT} 
& \multicolumn{3}{c}{DPO} \\
\cmidrule(lr){4-6} \cmidrule(lr){7-9}
Model &  &  & Iter 1 & Iter 2 & Iter 3 & Iter 1 & Iter 2 & Iter 3 \\
\midrule
Qwen2.5-7B\dag 
& 0.81 
& 0.82$_{+0.01\uparrow}$ 
& 0.82$_{+0.01\uparrow}$ 
& 0.82$_{+0.01\uparrow}$ 
& 0.82$_{+0.01\uparrow}$ 
& 0.82$_{+0.01\uparrow}$ 
& \underline{0.85}$_{+0.04\uparrow}$ 
& \textbf{0.86}$_{+0.05\uparrow}$ \\

Mistral-7B 
& 0.52 
& 0.48$_{-0.04\downarrow}$ 
& 0.48$_{-0.04\downarrow}$ 
& 0.48$_{-0.04\downarrow}$ 
& 0.48$_{-0.04\downarrow}$ 
& \textbf{0.52}\hspace{0.6em}$_{0.00}$\hspace{0.4em} 
& 0.48$_{-0.04\downarrow}$ 
& \underline{0.55}$_{+0.03\uparrow}$ \\

Llama-3.1-8B 
& 0.64 
& \underline{0.60}$_{-0.04\downarrow}$ 
& \textbf{0.61}$_{-0.03\downarrow}$ 
& \textbf{0.61}$_{-0.03\downarrow}$ 
& \underline{0.60}$_{-0.04\downarrow}$ 
& 0.51$_{-0.13\downarrow}$ 
& 0.43$_{-0.21\downarrow}$ 
& 0.48$_{-0.16\downarrow}$ \\
\bottomrule
\end{tabular}
\end{adjustbox}
\end{table*}

As intended, the target model \textbf{Qwen2.5-7B} shows a consistent decrease in performance throughout all stages, reaching its lowest value at the final DPO iteration (\textbf{0.71}, $\Delta=-0.10$). In contrast, the non-target models remain comparatively stable, with only marginal fluctuations ($\leq 0.02\Delta$). Notably, DPO training amplifies the divergence between the target and non-target models more cleanly than SFT, which plateaus after the first iteration. This suggests that preference-based fine-tuning more effectively reinforces the targeted degradation behaviour while preserving the performance of unaffected models. The results confirm that the paraphrase-sampling approach provides a viable mechanism for selective degradation without broad collateral effects.

Our other experiments show that loosening non-target models stability constraint leads to more significant score degradation in non-target model scores without notable change in target model quality. Loose stability constraint also leads to the generated paraphrases diverging from the original questions, often losing semantic alignment.

\paragraph{Maximisation of target model score}

We have performed the surrogate training experiment while aiming to maximize the target model score. The results are shown in Table \ref{tab:paraphrase_results_higher_target}. The target model score could be increased with DPO only by setting the target model weight in cost function \ref{par:cost_function} to 0.95.  Therefore, such training significantly impacts the other models' performance.

\section{Discussion}

By adding selective constraints to TextAttack and using a surrogate-model approach, we generated perturbations that caused one model to fail while leaving others unaffected. This reveals a robustness issue in current benchmarks: small, targeted changes can drastically reduce a specific model’s performance, undermining the reliability of benchmark comparisons. The selectivity of these attacks exposes deep differences in models’ inductive biases: despite similar overall accuracy, their internal representations and reasoning may diverge due to variations in tokenization, linguistic exposure, or reliance on surface cues over semantics.

The analysis of perturbed questions revealed several consistent linguistic and semantic patterns that account for the degradation in performance of the target model while the other ones remained unaffected. Firstly, many perturbations inserted low-frequency or contextually atypical tokens (e.g., “altitude”) into technical contexts as well as slight anomalies in phrasing or of the answer word field (e.g., “Answer:” → “note”). Some instruction-tuned models are tightly constrained by alignment, whereas others rely on common correct patterns, yielding consistent outputs.

\section*{Conclusion}

This study systematically investigates selective adversarial attacks on LLM benchmarks. In a white-box setting, we show that small, semantically valid perturbations can sharply degrade one model’s performance while leaving others intact. Using TextAttack with custom constraints and surrogate models, we established a reproducible framework for evaluating attack selectivity on MMLU. Results reveal serious weaknesses in benchmark evaluations: word-level changes—especially from the BAE attack—can invert model rankings, while character-level noise like DeepWordBug has little effect. Qwen models were most sensitive, making them useful for future study. These findings highlight benchmark fragility and the need for robustness analyses alongside leaderboard scores. Future work should pursue black-box attacks, cleaner benchmarks, and defense strategies. Beyond exposing risks, selective attacks may also guide positive optimization to improve alignment and reliability in LLMs.

\section*{Limitations}
To our knowledge, we are among the first to systematically examine the feasibility of selective adversarial attacks on large language models. Our study demonstrates that such attacks can be reproducibly constructed and can already produce substantial effects. Even modest changes in attack design can improve selectivity, though many questions remain—especially regarding black-box attacks, which may offer less mechanistic insight but greater realism in uncontrolled settings.

Another limitation concerns our benchmark choice. We used a subset of the MMLU benchmark rather than more specialized datasets. Because MMLU is widely used, parts of it may overlap with model pretraining data, reducing experimental purity. Future work should therefore employ cleaner, professionally curated benchmarks that better represent real-world conditions.

A key challenge ahead is developing defense mechanisms—both in models and benchmarks—to enhance robustness against adversarial perturbations. Selective attacks should be viewed not only as threats but also as opportunities: understanding how they alter model behavior can inform robustness training and targeted improvements in reasoning.

Finally, we plan to extend this work by designing explicitly selectivity-oriented attack algorithms. Achieving fine-grained control will require a deeper understanding of how linguistic cues, learning dynamics, and alignment objectives interact to make models differentially vulnerable.

\paragraph{Potential risks}

This work entails dual-use considerations. Insights into selective perturbations could be used to game leaderboards or bias evaluations against specific systems, undermining trust in benchmarks and distorting policy or procurement decisions. Evaluation overfitting is also a concern, as models may be tuned to known perturbations, reducing out-of-distribution robustness. Mitigations include staged disclosure of artifacts, perturbation-aware reporting with uncertainty intervals, independent audits with versioned benchmark governance, and ensemble/adaptive evaluations with randomized, regularly refreshed item pools.

% \section*{Acknowledgments}

\bibliography{custom}

\appendix

\section*{Appendix}
\label{sec:appendix}

\section{Surrogate model experiments}
\subsection{Models}
Our experimental framework involves multiple large language models:

%\begin{itemize}
%    \item \textbf{Target Model}: Qwen3-8B-Instruct
%    \item \textbf{Surrogate Model}: Qwen3-8B-Instruct (same as target)
%    \item \textbf{Comparison Models}: 
%    \begin{itemize}
%        \item Mistral-7B-Instruct 
%        \item Llama-3.1-8B-Instruct 
%    \end{itemize}
%\end{itemize}

\begin{table}[h]
\centering
\caption{Models evaluated in our experiments.}
\label{tab:models}
\begin{adjustbox}{max width=.5\textwidth}
\small
\begin{tabular}{@{} l l @{}}
\toprule
\multicolumn{1}{c}{Role} & \multicolumn{1}{c}{Model(s)} \\
\midrule
Target Model & \texttt{Qwen3-8B-Instruct} \\
Surrogate Model & \texttt{Qwen3-8B-Instruct} \\
 (same as target) & \\
Comparison Models & \begin{tabular}[t]{@{}l@{}}%
\texttt{Mistral-7B-Instruct}\\
\texttt{Llama-3.1-8B-Instruct}
\end{tabular} \\
\bottomrule
\end{tabular}
\end{adjustbox}
\end{table}

\subsection{Scoring Strategy}
We use likelihood-based scoring on benchmark: the model's response is taken as the choice (A-D) with the highest softmax-normalized log-probability at the final token position.

\subsection{LoRA Configuration}

We employ Low-Rank Adaptation (LoRA) \cite{hu2022lora} for parameter-efficient fine-tuning with the following configuration:

%\begin{itemize}
%    \item \textbf{Rank ($r$)}: 256
%    \item \textbf{Alpha ($\alpha$)}: 256 (scaling factor = $\alpha/r = 1.0$)
%    \item \textbf{Target Modules}: Self-attention projections and MLP layers.
%    \item \textbf{Dropout}: 0.0 (no dropout applied)
%    \item \textbf{Bias}: None (bias parameters not trained)
%\end{itemize}

\begin{table}[h]
\centering
\caption{LoRA configuration used in our experiments.}
\label{tab:lora_hparams}
\begin{adjustbox}{max width=.5\textwidth}
\small
\begin{tabular}{@{} l l @{}}
\toprule
\multicolumn{1}{c}{Hyperparameter} & \multicolumn{1}{c}{Value} \\
\midrule
Rank ($r$) & 256 \\
Alpha ($\alpha$) & 256 (scaling factor $\alpha/r = 1.0$) \\
Target Modules & Self-attention projections and MLP layers \\
Dropout & 0.0 (no dropout applied) \\
Bias & None (bias parameters not trained) \\
\bottomrule
\end{tabular}
\end{adjustbox}
\end{table}

\subsection{Training Procedure}
We utilize Direct Preference Optimization (DPO) as our primary training method.

%\paragraph{Hyperparameters}

%\begin{itemize}
%    \item \textbf{Batch Size}: 1 per device
%    \item \textbf{Number of Paraphrases for each Question}: 5
%    \item \textbf{Learning Rate}: $5\mathrm{e}{-5}$, constant
%    \item \textbf{Number of Training Epochs}: 3
%    \item \textbf{Optimizer}: Adam
%    \item \textbf{Gradient Accumulation Steps}: 4
%    \item \textbf{DPO $\beta$}: 0.1
%\end{itemize}

\begin{table}[h]
\centering
\caption{Training and optimization hyperparameters used in our experiments.}
\label{tab:train_hparams}
\begin{adjustbox}{max width=.5\textwidth}
\small
\begin{tabular}{@{} l l @{}}
\toprule
\multicolumn{1}{c}{Hyperparameter} & \multicolumn{1}{c}{Value} \\
\midrule
Batch size & 1 per device \\
Number of paraphrases for each question & 5 \\
Learning rate & $5\mathrm{e}{-5}$, constant \\
Number of training epochs & 3 \\
Optimizer & Adam \\
Gradient accumulation steps & 4 \\
DPO $\beta$ & 0.1 \\
\bottomrule
\end{tabular}
\end{adjustbox}
\end{table}

\subsection{Cost Function}
\label{par:cost_function}
Our objective for paraphrase assessment incorporates a balanced cost function with equal weighting:

\begin{equation}
    \mathcal{L} = 0.5 \cdot \mathcal{L}_{\text{target}} + 0.5 \cdot \mathcal{L}_{\text{other}}
\end{equation}

with $\mathcal{L}_{\text{target}}$ minimized if the target models answers all the answers wrong and $\mathcal{L}_{\text{other}}$ minimized when all the other models do not change their answers on paraphrased question.

\subsection{Hardware and Computational Resources}

%\begin{itemize}
%    \item \textbf{GPU:} 1$\times$ NVIDIA H100 SXM5 80GB
%    \item \textbf{CPU:} AMD EPYC-Milan (180) @ 3.599GHz
%    \item \textbf{RAM:} 1.45T
%    \item \textbf{Training time per cycle: } 8 min
%    \item \textbf{Total compute hours: } 5h
%\end{itemize}

\begin{table}[h]
\centering
\caption{Hardware and computational resources used in our experiments.}
\label{tab:hardware_compute}
\begin{adjustbox}{max width=.5\textwidth}
\small
\begin{tabular}{@{} l l @{}}
\toprule
\multicolumn{1}{c}{Resource} & \multicolumn{1}{c}{Specification} \\
\midrule
GPU & 1$\times$ NVIDIA H100 SXM5 80GB \\
CPU & AMD EPYC-Milan (180) @ 3.599GHz \\
RAM & 1.45T \\
Training time per cycle & 8 min \\
Total compute hours & 5 h \\
\bottomrule
\end{tabular}
\end{adjustbox}
\end{table}

\section{Surrogate model prompting}

Here we list the prompts used for generating paraphrased questions and benchmarking the models.

\paragraph{Paraphrase request}

\noindent The following system prompt was used during paraphrase generation. The user message contains the initial question.

\begin{tcolorbox}[colback=gray!5,colframe=gray!40,arc=0pt,boxrule=0.5pt]
\small
\texttt{%
You are a helpful assistant specialized in rephrasing academic and scientific questions.\\
When given a question or prompt, your task is to produce a new version that preserves its original meaning while changing the wording and phrasing naturally.\\
\\
The rephrased question should remain semantically equivalent, grammatically correct, and appropriate for domains such as science, mathematics, engineering, social sciences, or the humanities (as represented in the MMLU benchmark).\\
\\
Do not answer the question --- only rephrase it clearly and precisely.
}
\end{tcolorbox}

\paragraph{Evaluation Prompt}

\noindent The following system prompt was used during evaluation to guide the model’s multiple-choice answering behavior:

\begin{tcolorbox}[colback=gray!5,colframe=gray!40,arc=0pt,boxrule=0.5pt]
\small
\texttt{%
You are a helpful assistant that answers multiple-choice questions.\\
For each question, choose the single best answer from the provided options (A, B, C, D, etc.).\\
Respond using only the letter corresponding to your selected answer.
}
\end{tcolorbox}

\section{Surrogate model paraphrase results}

\noindent Below is an example showing the original MMLU question and several paraphrased versions generated by our surrogate model.

\begin{tcolorbox}[colback=gray!5,colframe=gray!40,arc=0pt,boxrule=0.5pt]
\small
\textbf{Initial Question:}\\
Find all $c \in \mathbb{Z}_3$ such that $\mathbb{Z}_3[x]/(x^2 + c)$ is a field.\\[6pt]

\textbf{Paraphrased Questions:}
\begin{enumerate}
    \item Determine all values of $c$ in $\mathbb{Z}_3$ for which the quotient ring $\mathbb{Z}_3[x]/(x^2 + c)$ forms a field.
    \item Determine all values of $c$ in $\mathbb{Z}_3$ for which the quotient ring $\mathbb{Z}_3[x]/(x^2 + c)$ forms a field.
    \item Determine all $c$ in $\mathbb{Z}_3$ for which the quotient ring $\mathbb{Z}_3[x]/(x^2 + c)$ forms a field.
    \item Determine all values of $c$ in $\mathbb{Z}_3$ for which the quotient ring $\mathbb{Z}_3[x]/(x^2 + c)$ forms a field.
    \item Determine all $c$ in $\mathbb{Z}_3$ for which the quotient ring $\mathbb{Z}_3[x]/(x^2 + c)$ forms a field.
\end{enumerate}
\end{tcolorbox}

\begin{tcolorbox}[colback=gray!5,colframe=gray!40,arc=0pt,boxrule=0.5pt]
\small
\textbf{Initial Question:}\\
Which of the following is an example of the use of a device on the Internet of Things (IoT) ?\\[6pt]

\textbf{Paraphrased Questions:}
\begin{enumerate}
    \item Which device below is an example of Internet of Things (IoT) technology in use?
    \item Which example below illustrates the application of a device in the Internet of Things (IoT)?
    \item Which example below demonstrates the application of a device within the Internet of Things (IoT) ecosystem?
    \item Which device below exemplifies the application of the Internet of Things (IoT)?
    \item Which example below demonstrates the application of a device in the Internet of Things (IoT)?
\end{enumerate}
\end{tcolorbox}

\begin{tcolorbox}[colback=gray!5,colframe=gray!40,arc=0pt,boxrule=0.5pt]
\small
\textbf{Initial Question:}\\
Large triplet repeat expansions can be detected by\\[6pt]

\textbf{Paraphrased Questions:}
\begin{enumerate}
    \item How can large triplet repeat expansions be identified?
    \item How can large triplet repeat expansions be identified?
    \item How can large triplet repeat expansions be identified?
    \item What methods can be used to identify large triplet repeat expansions?
    \item How can large triplet repeat expansions be identified?
\end{enumerate}
\end{tcolorbox}

\end{document}